\documentclass[lettersize,journal]{IEEEtran}
\usepackage{amsmath,amsfonts}
\usepackage{algorithmic}
\usepackage{algorithm}
\usepackage{array}
\usepackage[caption=false,font=normalsize,labelfont=sf,textfont=sf]{subfig}
\usepackage{textcomp}
\usepackage{stfloats}
\usepackage{url}
\usepackage{verbatim}
\usepackage{graphicx}
\usepackage[switch]{lineno}
\usepackage{cite}
\usepackage{subfig}
\usepackage{pifont}
\usepackage{bbding}
\usepackage{multirow}
\usepackage{makecell}
\usepackage{booktabs}%
\begin{document}

\title{Dual-Correlation Hypergraph Network for Unaligned RGBT Video Object Detection and A Large-scale Benchmark}

\author{Qishun Wang, Yapeng Li, Zhengzheng Tu, Chenglong Li,~\IEEEmembership{Senior Member,~IEEE, and Bin Luo}
        
\thanks{Corresponding author: Zhengzheng Tu.}

\thanks{Qishun Wang, Yapeng Li, Zhengzheng Tu, and Bin Luo are with Anhui Provincial Key Laboratory of Multimodal Cognitive Computation, School of Computer Science and Technology, Anhui University, Hefei 230601, China (e-mail: qishunahu@163.com; liyapeng\_w@163.com; zhengzhengahu@163.com; luobin@ahu.edu.cn). Chenglong Li is with Anhui Provincial Key Laboratory of Multimodal Cognitive Computation, School of Artificial Intelligence, Anhui University, Hefei, 230601, China (e-mail: lcl1314@foxmail.com).}}

\markboth{Journal of \LaTeX\ Class Files,~Vol.~14, No.~8, August~2021}%
{Shell \MakeLowercase{\textit{et al.}}: A Sample Article Using IEEEtran.cls for IEEE Journals}


\maketitle

\begin{abstract}
RGB-Thermal (RGBT) Video Object Detection (VOD) has gained significant attention because of the limitations of conventional RGB-based VOD methods under challenging conditions, such as low light, heavy fog, and adverse weather, etc. However, spatial misalignment commonly exists between RGBT image pairs. To address this, we propose a Dual-Correlation Hypergraph Network (DCHNet) that captures high-dimensional complementary information by explicitly modeling two types of correlations: temporal correlation across consecutive frames and spatial correlation from cross-modal features. Specifically, we first design a Patch-based Spatial Alignment Module (PSAM) to sequentially align the multimodal features at the local region level. Subsequently, we propose a Dual Hypergraph Fusion Module (DHFM), which constructs temporal and multimodal hypergraphs, respectively, to enhance object characteristic through dual-correlation learning. Furthermore, the field currently lacks a large-scale, scene-diverse benchmark dataset for comprehensive evaluation. Therefore, we construct DVT-VOD1000, a large-scale RGBT VOD dataset containing 1,000 video sequences with 103,464 RGBT image pairs. The dataset covers diverse scenarios, including campuses, parks, traffics, rural areas, night scenes, rainy weather, and snowy weather. Comprehensive experiments on VT-VOD50 and our DVT-VOD1000 demonstrate that DCHNet achieves state-of-the-art detection accuracy. The dataset and source code will be made publicly available on \url{https://github.com/tzz-ahu/} to support academic research.  
\end{abstract}


\section{Introduction}
\IEEEPARstart{A}{s} a key visual perception task, Video Object Detection (VOD) serves as a core component in fields like security surveillance and autonomous driving \cite{huang2013highly,taylor2016anomaly,li2019aads}. However, VOD that relies solely on RGB information remains susceptible to robustness issues in extreme scenarios, such as low-light conditions at night, overexposure, and adverse weather including rain, snow, and fog. To mitigate these limitations, the work by Wang et al. \cite{wang2025erasure} introduces a method that combines RGB and thermal (RGBT) for VOD and creates a corresponding benchmark dataset, VT-VOD50. The authors apply manual processing to this dataset to achieve consistency in both resolution and spatial distribution. However, the VT-VOD50 dataset contains only 100 RGBT video sequences and fewer than 10,000 image pairs. Furthermore, its data originates exclusively from traffic scenarios. This limitation imposes a constraint on the comprehensive evaluation of RGBT VOD methods and adversely affects model generalization.

\begin{figure}[t]
        \centering
\includegraphics[width=0.96\linewidth]{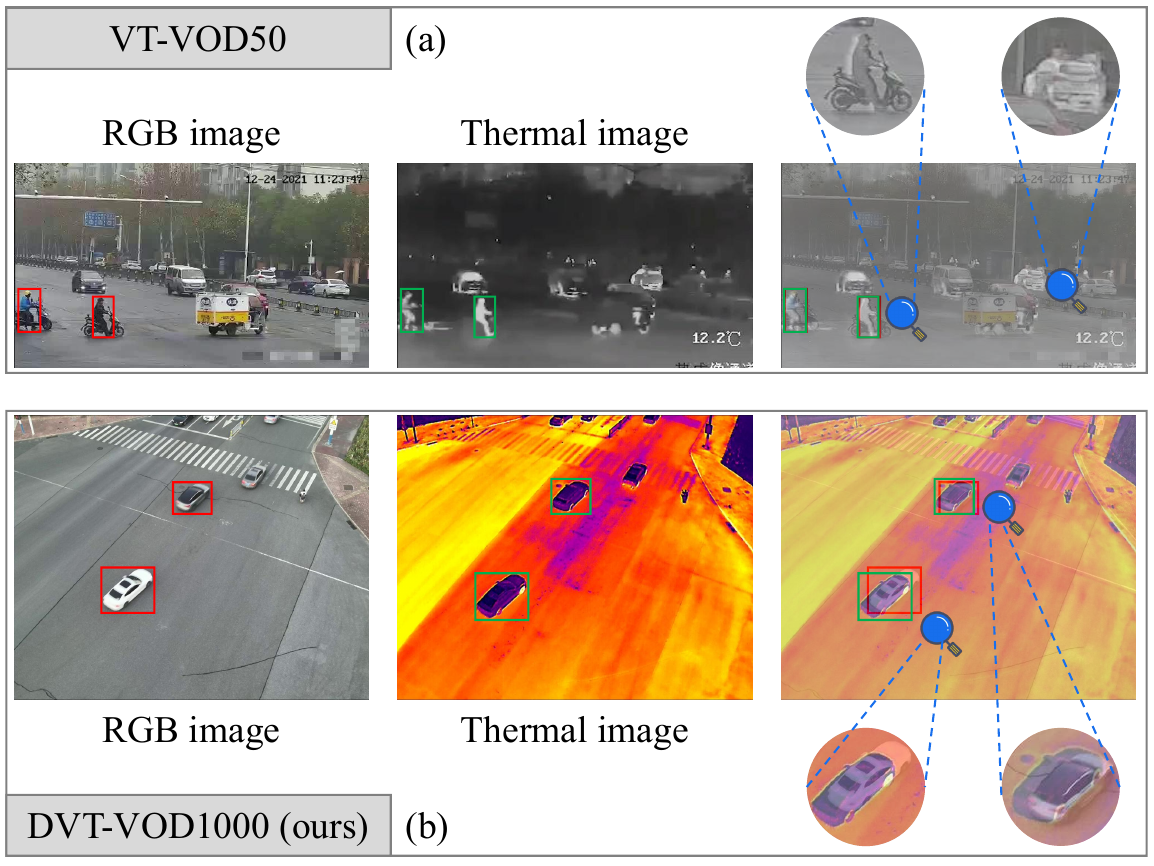}
        \caption{Existing VT-VOD50 (a) and our proposed DVT-VOD1000 (b), commonly exhibit spatial misalignment of objects. As shown in (b), this issue manifests as some regional variations in the degree of misalignment. The third column shows the result of overlaying the RGB and thermal images—each adjusted to 50\% transparency—to clearly indicate the object location.}
        \label{motivation}
    \end{figure}

Wang et al. \cite{wang2025erasure} introduce EINet for RGBT VOD, which employs a negative activation function to identify noise regions in thermal images and suppress background noise in RGB features. PTMNet \cite{wang2025high} combines early and middle fusion strategies to enhance detection accuracy, where early fusion reduces the semantic gap between modalities. Despite their effectiveness, both methods operate under a strong assumption: that the RGB and thermal inputs are pixel‑wise aligned. This assumption, however, is frequently violated in practice. As illustrated in Fig. \ref{motivation} (a), even in the manually aligned VT‑VOD50 dataset, noticeable spatial offsets persist between the two modalities. These misalignments pose a fundamental challenge to existing fusion paradigms: pixel‑wise fusion operations force the network to combine features that correspond to different spatial locations, which not only weakens the object feature but also introduces irrelevant background information, ultimately degrading detection performance. Some recent works attempt to address this issue by incorporating extra alignment modules to pre‑align the modality inputs \cite{wang2025mixturescaleexpertsalignmentfree,wang2025multimodal}. However, such strategies inevitably introduce additional computational overhead and are sensitive to alignment errors. Moreover, they treat misalignment as a preprocessing problem rather than rethinking the fusion mechanism itself.

In summary, there are two main issues: first, existing methods neglect the misalignment problem that is commonly present in RGBT video pairs; second, the RGBT VOD field lacks large‑scale, multi‑scene benchmark datasets. To address these issues, we first propose a novel Dual-Correlation Hypergraph Network (DCHNet) for RGBT VOD that effectively mitigates the misalignment challenge through a patch-by-patch alignment strategy. Specifically, we first propose a Patch-based Spatial Alignment Module (PSAM), which addresses the problem of regionally varying alignment by applying decentralized local spatial transformations. We then utilize Local Binary Pattern (LBP) \cite{ojala1994performance} to extract global detailed positional information from RGB features and inject it into the thermal features to achieve further global alignment. Finally, we propose a Dual-Correlation Hypergraph Fusion Module (DHFM) that models the correlation between RGB and thermal images, as well as the correlation between RGB and its temporally adjacent frame, enabling spatial-temporal feature fusion. DCHNet avoids doing the explicit spatial alignment by constructing dual‑correlation hypergraphs. Instead of forcing pixel‑level correspondences, the hypergraphs capture high‑order semantic relationships across modalities at a coarse‑grained level. Furthermore, we construct a large-scale evaluation dataset for RGBT VOD, named DVT-VOD1000. The dataset covers a wide range of spatial scenarios—such as campuses, rural areas, transportation zones, and suburban areas—as well as diverse time scenarios, including nighttime, rainy days, and snowy weather. DVT-VOD1000 contains 1,000 RGBT videos and over 103,000 RGBT image pairs. As shown in Fig. \ref{motivation} (b), overlaying the RGB and thermal images from DVT-VOD1000 (third column) reveals spatial misalignment between the two modalities.

The main contributions of this work are summarized as follows:
\begin{itemize}
    \item We propose a DCHNet that alleviates the prevalent misalignment problem in RGBT-VOD by integrating local spatial alignment with global positional priors, and subsequently learns high-order cross-modal correlations and performs feature fusion by constructing a hypergraph between multimodal features.
    \item We propose PSAM to perform patch‑wise spatial transformations on unaligned RGBT image pairs for initial alignment. Next, we introduce DHFM, which employs hypergraphs to jointly model higher‑order correlations among RGB images, thermal images, and temporal frames, thereby achieving spatial-temporal feature fusion.
    \item We construct DVT-VOD1000, a large-scale benchmark dataset for RGBT VOD. It captures diverse real-world spatial scenes using drone platforms, comprising 1,000 aligned RGB-thermal video sequences and over 103,000 RGBT image pairs.
    \item Evaluations across a wide range of detectors on our DVT-VOD1000 and the VT-VOD50 indicate that DCHNet achieves state-of-the-art performance. Extensive experiments further prove its effectiveness in handling unaligned scenarios.
\end{itemize}

\section{Related Work} 
\subsection{Multispectral Object Detection}
Multispectral object detection addresses the limitations of RGB-based detection under challenging imaging conditions, such as extreme illumination, rain, and fog \cite{zhou2025cofnet,chen2025amfd,shin2025ssmpd,yang2026dwsf}.

Fang et al. \cite{qingyun2021cross} introduce a cross-modal fusion transformer that enhances the capacity to model long-range dependencies and global information during multimodal fusion. They argue that its attention mechanism uncovers latent relationships between RGB and thermal features while also capturing intra-modal contextual information. Similarly, Shen et al. \cite{shen2024icafusion} propose a novel dual cross-attention feature fusion method that captures complementary information between modalities by means of cross-modal queries. Zhao et al. \cite{zhao2025removal} propose a coarse-to-fine approach for multimodal fusion. Specifically, they first design a redundant spectrum removal module to perform coarse denoising within each modality, and then propose a dynamic feature selection module that adaptively selects and fuses complementary features. Hu et al. \cite{hu2025ei} propose to leverage light intensity for dynamically analyzing the distinct contributions of visible and infrared data, and then utilize the edge information for guiding the enhanced network to locate objects during the fusion process.

To address the issue that CNNs focus only on local features, while Transformers are overly complex, Dong et al. \cite{dong2025fusion} propose using a state space model \cite{liu2024vmamba} to first map RGBT features to a hidden state space, thereby reducing modal differences before performing efficient fusion. To tackle the spatial misalignment in RGBT image pairs, Zhao et al. propose RGFNet \cite{zhao2025reflectance}, which leverages reflection components to progressively register RGB and thermal images for mitigating the challenge under varying lighting conditions. In contrast, our design makes our approach inherently robust to spatial misalignments, as it relies on holistic context and relational reasoning rather than strict spatial correspondence.

\subsection{Video Object Detection}
VOD extends image-based detection by incorporating temporal information of video sequences, thereby mitigating challenges such as object blurring and occlusion that often arise in individual frames.

Zhu et al. \cite{zhu2017deep} argue that extracting features for every frame in a video is computationally expensive; therefore, they propose to employ optical flow to propagate feature maps across adjacent frames, thereby reducing redundant computation. Similarly, Zhu et al. \cite{zhu2017flow} propose to leverage optical flow to deform and aggregate features from neighboring frames into the current frame, thereby enhancing temporal feature representation. To address the issue that objects are continuously occluded in video sequences, Chen et al. \cite{chen2020memory} propose to synergistically enhance object features in the current frame by combining local positional correlation with global semantic similarity for effectively mitigating occlusion or blurring. RGB-based VOD exhibits limited robustness under extreme imaging conditions such as low-light environments. To address this, Wang et al. \cite{wang2025erasure} first introduce the RGBT VOD task and propose a method that leverages thermal features after negative activation to suppress noise in RGB features for enhancing foreground object representation. To mitigate domain differences in multimodal fusion, Wang et al. propose PTMNet \cite{wang2025high}, which integrates early and middle fusion strategies to enhance fusion performance. Specifically, PTMNet performs pixel-level fusion of different modalities at the network input stage, thereby reducing semantic differences between modalities during feature extraction.

\begin{figure*}[t]
        \centering
\includegraphics[width=0.98\linewidth]{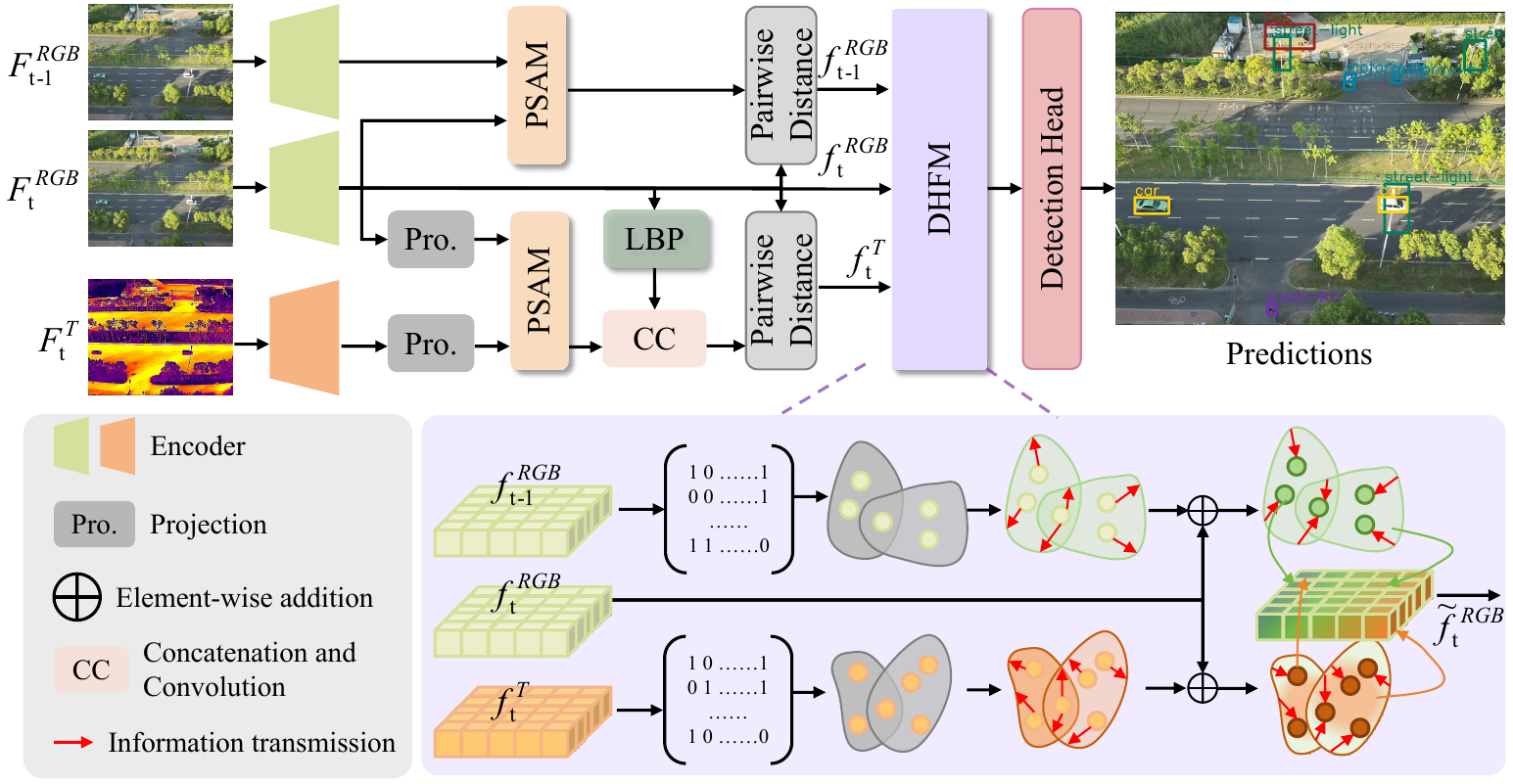}
        \caption{The core architecture of DCHNet. The PSAM first performs local spatial alignment on the two input feature maps in a region-by-region manner. Subsequently, we utilize LBP to extract global object location information from $F_{t}^{RGB}$, which is injected into $F_{t}^{T}$ to achieve global calibration. Following this, the proposed DHFM constructs high-order correlations among temporally adjacent and cross-modal features via hypergraph learning.}
        \label{network}
    \end{figure*}

\section{Method}
\subsection{Overview}
Our DCHNet (as shown in Fig. \ref{network}) builds upon the YOLOV8 framework \cite{Jocher_Ultralytics_YOLO_2023}, adopting its feature encoder and detection head. The DCHNet employs a shared backbone network for two RGB frames $F_{t}^{RGB}$ and $F_{t-1}^{RGB}$, whereas the thermal frame $F_{t}^{T}$ uses a separate backbone that shares the same structure but maintains independent parameters. Next, the PSAM performs patch-level spatial transformation alignment on both temporally adjacent features and multimodal features. Given that spatial discrepancies caused by multimodal sensor distortion are more substantial, we employ LBP to extract global position details from the current RGB frame, which are then injected into the corresponding thermal frame. Subsequently, DHFM conducts hypergraph-based feature fusion on the three previously derived features ($f_{t-1}^{RGB}$, $f_{t}^{RGB}$ and $f^{T}_{t}$). The fused representation is finally fed into the detection head to generate the prediction result. DCHNet adopts the same loss function as the baseline, and its detailed formulation is omitted for brevity.

\subsection{PSAM: Patch-based Spatial Alignment Module}
\begin{figure*}[t]
        \centering
\includegraphics[width=0.98\linewidth]{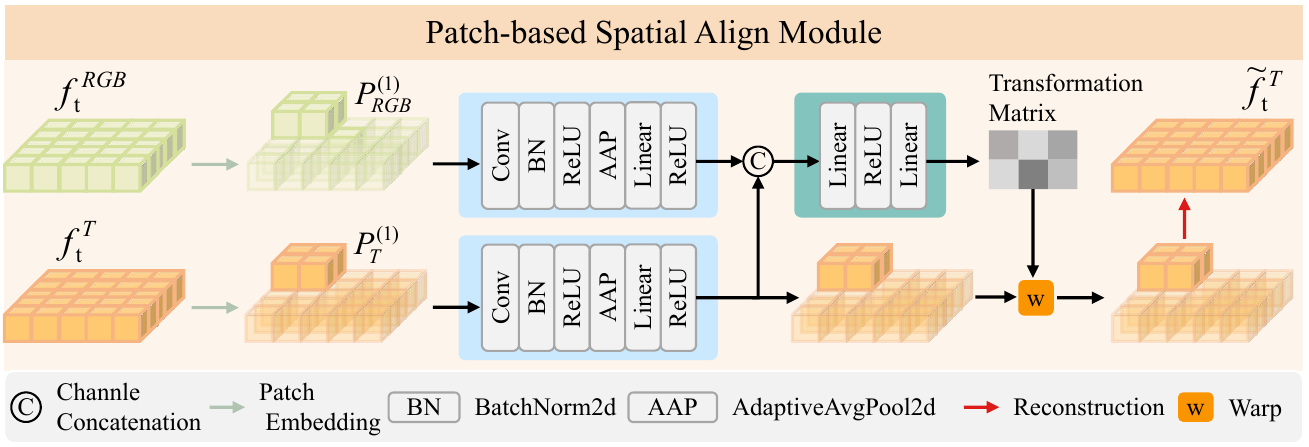}
        \caption{Illustration of the patch-based alignment process in the PSAM.}
        \label{psam}
    \end{figure*}

Misalignment is a common challenge in RGBT VOD datasets. The proposed DVT-VOD1000 benchmark dataset provides varying degrees of misalignment across different spatial regions. The region inconsistency motivates the implementation of localized spatial transformations on the thermal image to achieve effective spatial alignment. To overcome this challenge, we introduce the PSAM, whose architecture is depicted in Fig. \ref{psam}.
The PSAM takes two input features, denoted as $f_{t}^{RGB} \in \mathbb{R}^{C \times H \times W}$ and $f_{t}^{T} \in \mathbb{R}^{C \times H \times W}$. It begins by dividing each feature into multiple non-overlapping patches in the spatial dimension using a sliding window, thereby generating two corresponding patch sequences $\left\{P^{(i)}_{\mathrm{RGB}}\right\}_{i=1}^N$ and $\left\{P^{(i)}_{\mathrm{T}}\right\}_{i=1}^N$. To alleviate the computational cost brought by aligning multiple patch blocks, we first employ a 1×1 convolutional layer to reduce the channel dimensionality. The resulting feature block is then transformed into a compact descriptor through global average pooling and flattening. Finally, this descriptor is passed through a linear layer followed by a non-linear activation function to produce discriminative features. This procedure is summarized by the following equation:
 \begin{equation}
        f_{RGB}^{(i)} = \phi(P_{RGB}^{(i)}),  f_{T}^{(i)} = \phi(P_{T}^{(i)}). 
    \label{}
    \end{equation}
Here, $\phi()$ denotes a feature encoder with shared weight parameters. The pairs of patch features are subsequently concatenated along the channel dimension and fed into the affine transformation prediction, as follows:
\begin{equation}
    \vartheta^{(i)} = \psi([f_{RGB}^{(i)},f_{T}^{(i)}])\in \mathbb{R}^6,
\end{equation}
where $\psi$ corresponds to the affine transformation prediction process. The $\vartheta^{(i)}$ is subsequently transformed into an affine matrix $A^{(i)}$ according to the following formula:
\begin{equation}
        A^{(i)} = \left[ \begin{array}{ccc}
1+\vartheta^{(i)}_1 & \vartheta^{(i)}_2 & \vartheta^{(i)}_3 \\
\vartheta^{(i)}_4 & 1+\vartheta^{(i)}_5 & \vartheta^{(i)}_6
\end{array} \right].
\end{equation}
The sampling grid generated by matrix $A^{(i)}$ is used to spatially transform $P_{T}^{(i)}$, aligning it with the $P_{RGB}^{(i)}$. The transformed $P_{T}^{(i)}$ is then re-inserted into the thermal feature $f^{T}_t$ at its original spatial indices. By iterating this procedure, the aligned feature $\tilde{f}^{T}_t$ consistent with the RGB feature $f^{RGB}_t$ is reconstructed. The procedure for aligning the adjacent temporal frame using PSAM is similar to the one described above; it simply requires replacing input $f^{T}_t$ with $f^{RGB}_{t-1}$.

\subsection{DHFM: Dual-Correlation Hypergraph Fusion Module}
Most existing RGBT VOD methods are limited to modeling pairwise feature relationships, and consequently cannot simultaneously capture multimodal and temporal consistency. To address this, we propose the DHFM that models spatiotemporal feature consistency in a unified manner and effectively captures higher-order dependencies.

The schematic of the DHFM is depicted in Fig. \ref{network}. It comprises two symmetrical branches with identical structures and shared weights, dedicated to modeling temporal information and capturing multimodal complementary features, respectively. The module takes three feature maps ($f^{RGB}_{t-1} \in \mathbb{R}^{C \times H \times W}$, $f^{RGB}_t \in \mathbb{R}^{C \times H \times W}$ and $f^{T}_t \in \mathbb{R}^{C \times H \times W}$) as inputs, which correspond to middle stage features of $F^{RGB}_{t-1}$, $F^{RGB}_t$ and $F^{T}_t$. The DHFM module first reshapes these input features into a set of vertex representations:
\begin{equation}
    [\left\{V_{t-1}^{RGB}\right\}^{N}_{i=1},\left\{V_{t}^{RGB}\right\}^{N}_{j=1},\left\{V_{t}^{T}\right\}^{N}_{k=1}] \in \mathbb{R}^{N \times C},N=H \times W.
\end{equation}
We take the multimodal branch as an example to explain this process. For the given pair of nodes ($\left\{V_{t}^{RGB}\right\}_{i}$, $\left\{V_{t}^{T}\right\}_{j}$), we compute the Euclidean distance $d_{ij}$ between them. Subsequently, a threshold $\tau$ (set to 8 in this work) is applied to filter out low-relevance edges with distances exceeding this value, thereby generating the final hyperedges. This process finally produces a binary sparse matrix $H\in \mathbb{R}^{N \times N}$, which stores the hyperedges for all node pairs. A two-step message aggregation is then performed based on the correlation matrix $H$. Based on the correlation matrix $H$, we perform a two‑step message aggregation between the two modalities. In the first step, we aggregate vertex information from the thermal features onto the hyperedges. Specifically, each hyperedge integrates the features of all thermal vertices that are connected to it, with the aggregation weights determined by the correlation values in $H$. This step produces a set of hyperedge representations that encode cross‑modal relationships from the thermal perspective. In the second step, we propagate the information from the hyperedges back to the RGB vertices. For each RGB vertex, it collects the aggregated messages from all hyperedges that connect to it, again weighted by the corresponding correlation values in $H$. Through this bidirectional aggregation process, the RGB features are progressively enhanced with cross‑modal information from the thermal modality, enabling effective semantic interaction and complementarity between the two modalities. These two steps are formulated as follows:
\begin{equation}
    \tilde{V}_t^{RGB} = \delta_2(\delta_1(V_t^{T},H^T),H).
\end{equation}
Here, $\delta_1$ and $\delta_2$ denote the message propagation from vertex to hyperedge and from hyperedge to vertex, respectively. For the detailed architectural designs, please refer to \cite{feng2024hyper}. Subsequently, the new vertex set $\tilde{V}_t^{RGB}$, which aggregates the thermal features is reconstructed back into a feature map, yielding the updated representation $\tilde{f}^{RGB}_t$.

Following the same procedure described above, the temporal information for features $f^{RGB}_t$ and $f^{RGB}_{t-1}$ is aggregated. In essence, the DHFM seamlessly integrates multimodal and temporal fusion for RGBT VOD within a unified hypergraph-based framework, as a holistic high-order information transfer process.

\section{DVT-VOD1000: A Large-scale Drone-based RGBT VOD Dataset}
\subsection{Motivation and Data Collection}
The field of RGBT VOD attracts growing research interest. However, the widely used VT-VOD50 \cite{wang2025erasure} dataset offers limited ability for comprehensively model evaluation due to insufficient data and scenarios. To address these limitations, this work introduces DVT-VOD1000, a new large-scale benchmark for RGBT VOD. With its extensive real-world scenes and substantial size, DVT-VOD1000 is a more robust platform for both training and evaluating video object detectors.

The DVT-VOD1000 is captured using a DJI Mavic 3 Enterprise series drone platform \cite{DJI_Mavic3_2025}, specifically the Mavic 3T. This platform integrates a wide-angle camera, a telephoto camera, and a thermal imaging camera in its gimbal, making it suitable for applications such as urban patrols, wilderness search and rescue, and nighttime operations. The wide-angle camera is equipped with a 48-megapixel sensor and provides an 84° field of view, supporting video recording at 1920 $\times$ 1080 resolution and 30 Frames Per Second (FPS). The telephoto camera features a 12-megapixel sensor with a 15° field of view. The thermal imaging camera supports a temperature measurement range of $-20$\,\textdegree C to 150\,\textdegree C and records video at a resolution of 640 $\times$ 512. Both RGB and thermal modalities are recorded simultaneously. The built-in algorithm crops the RGB image and applies super-resolution to the thermal image, resulting in a unified output resolution of 960 $\times$ 770 for both modalities. Finally, the RGB and thermal videos are segmented into multiple frames at 24 FPS and stored in the dataset.
\subsection{Scene and Environmental Diversity}
The collection of the DVT-VOD1000 dataset spans over twelve months. It encompasses a wide range of real-world scenarios for both model training and evaluation.

\begin{figure*}
        \centering
\includegraphics[width=\linewidth]{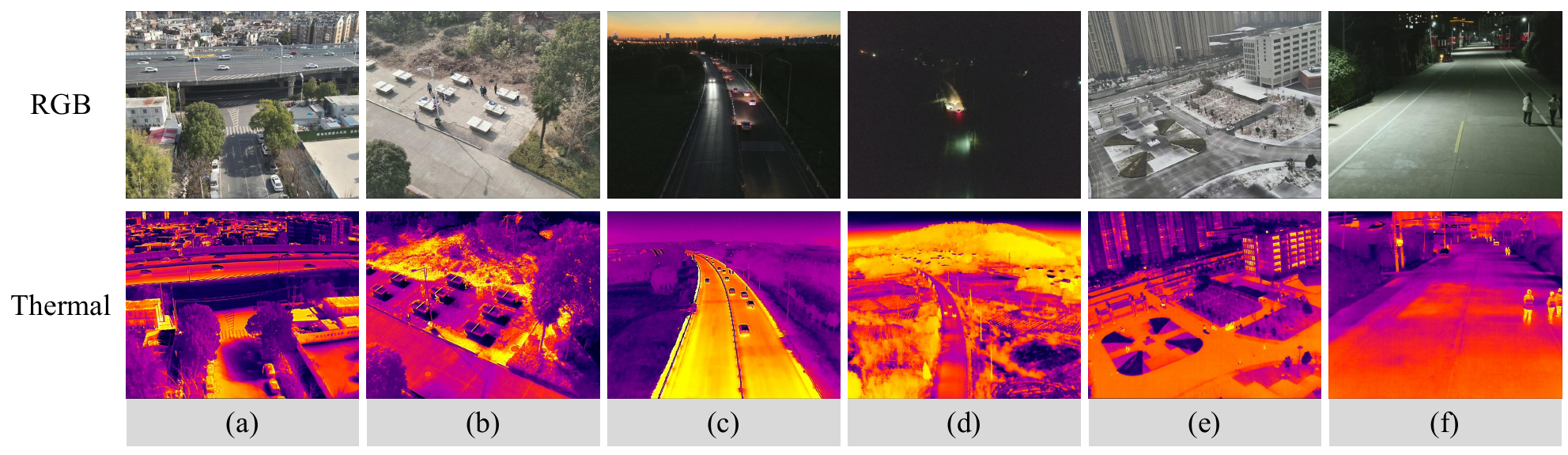}
        \caption{The DVT-VOD1000 dataset comprises a diverse range of RGBT image pairs captured across various typical scenarios.}
        \label{dataset_scene}
    \end{figure*}

To fully illustrate the scene diversity of the DVT-VOD1000 dataset, six representative RGBT image pairs are shown in Fig. \ref{dataset_scene}. Group (a) depicts a main urban road during peak hours, where intersecting roads and tree occlusion present challenges for object recognition. Group (b) presents the walking pedestrian scenario in a park, containing multiple object categories such as pedestrians, streetlights, and trash cans. There are also object occlusion and small objects. Group (c) shows a suburban road at dusk, where the dynamic range of the RGB image is significantly compressed and complicated by headlight glare. In contrast, the thermal image is clear. Group (d) captures a rural night scene, where ambient light is extremely low and RGB images suffer from significant noise for validating the model's detection capability under extreme low light, long distance, and small object conditions. Group (e) displays a snowy campus scene, where the thermal image enhances the contrast between foreground and background while the RGB image presents more complex background textures. Group (f) shows a low-altitude scene captured from a drone perspective similar to road surveillance, with object scale and viewing angle differing from those in previous groups.
    
\subsection{Annotation}
We adhere to the instance-level granularity standard for object detection tasks and expand the category definitions in three key aspects. The first is traffic participants, which include person, bicycle, car, taxi, motorcycle, bus, van, and truck. The second is traffic facilities, comprising traffic-light, zebra-crossing, road-sign, and traffic-sign. Finally is infrastructure element, which include trash-bin, street-light, and surveillance-camera. In total, the DVT-VOD1000 dataset encompasses 15 distinct categories.

We employ LabelImg as the annotation tool, which is a visual image annotation utility that supports exporting annotation files in PASCAL VOC \cite{everingham2010pascal} XML format and enables conversion to and from COCO \cite{lin2014microsoft} format. Each XML file corresponds to a single image frame and includes category labels and location information of all predefined objects present in the corresponding RGB image. Annotation is performed solely on the RGB modality to prevent potential conflicts arising from spatial misalignment between RGB and thermal images during network training. This approach ensures that the network optimizes its parameters based on annotations of RGB data. To ensure annotation quality and consistency, a team of over ten annotators label the data concurrently. A dedicated supervisor then checks all annotations.

\subsection{Statistics and Scale}
\begin{figure*}
        \centering
\includegraphics[width=\linewidth]{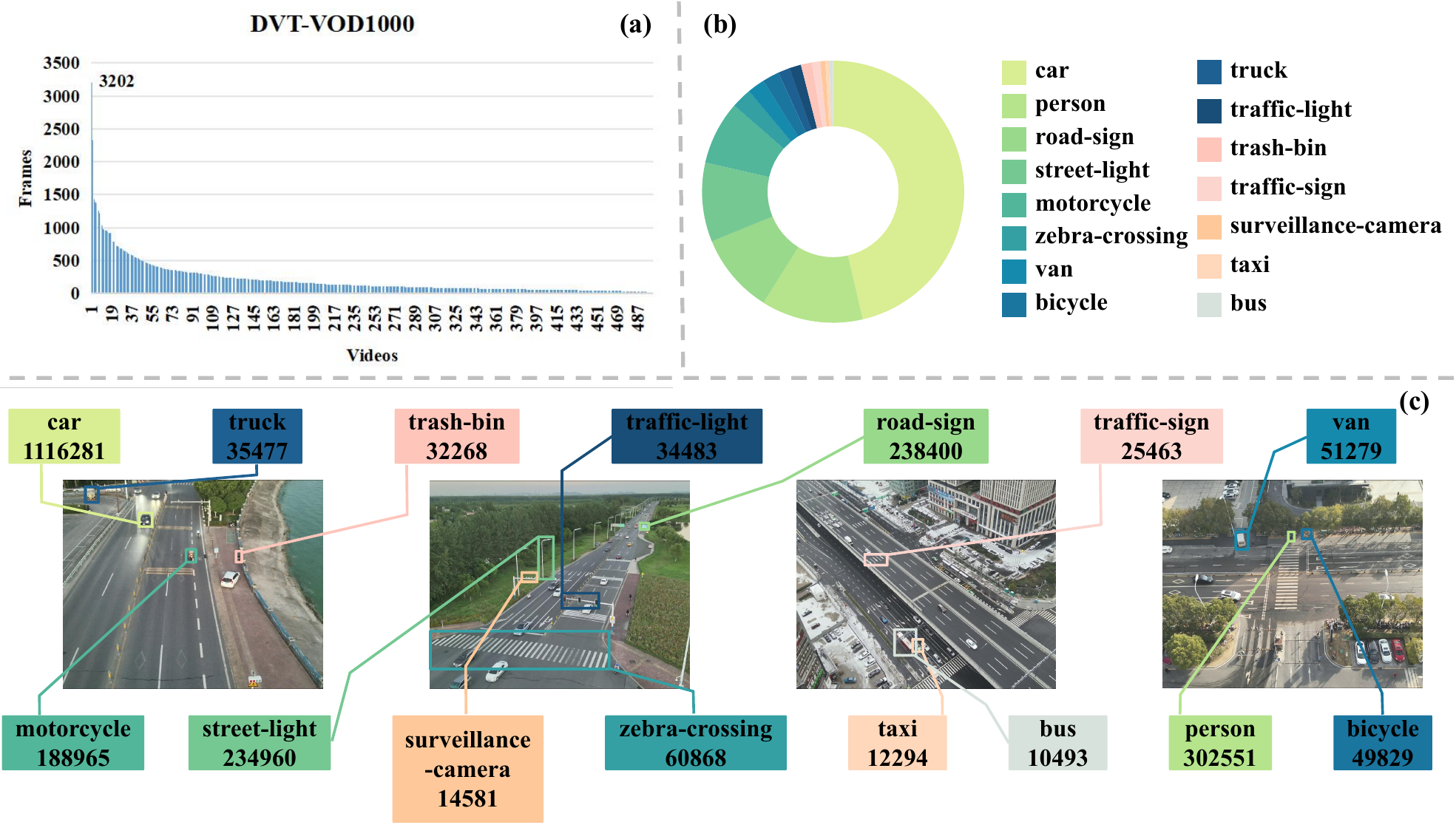}
        \caption{(a) depicts the video sequences in the DVT-VOD1000 dataset and their length distribution. (b) presents the distribution of the 15 object categories. (c) details the number of instances for each category.}
        \label{dataset_distribution}
    \end{figure*}

The DVT-VOD1000 dataset comprises 500 RGBT video pairs. The frame length distribution across these pairs is presented in Fig. \ref{dataset_distribution} (a), ranging from 15 to 3,202 frames, with the majority containing fewer than 1,500 frames (approximately one minute in duration).
Fig. \ref{dataset_distribution} (b) illustrates the category distribution and respective proportions of the 15 object classes in DVT-VOD1000. The ``car" class constitutes nearly half of all instances, followed by the ``person" class. Conversely, categories such as ``taxi" and ``bus" contain relatively few instances. This long-tailed distribution is consistent with real-world traffic scenarios, demonstrating that the dataset's statistical characteristics reflect object class imbalances in real-world.

Table \ref{dataset_compara} presents a comprehensive comparison of our DVT-VOD1000 with existing RGBT object detection datasets, including dataset size, resolution, and number of classes, etc. Existing datasets primarily consist of static images, where the largest two include KAIST (95,328 images) \cite{choi2018kaist} and DroneVehicle (56,878 images) \cite{sun2022drone}. In contrast, our DVT-VDO1000 is a video dataset comprising 206,928 frames—more than twice the size of KAIST. This substantial volume of sequential data provides rich temporal contextual information, being helpful for moving object detection and temporal consistency learning. The DVT-VOD1000 dataset has medium-to-high resolution, thereby avoiding the complexities associated with multi-scale training. While most existing datasets prioritize precise alignment, real-world RGB and thermal infrared sensors frequently exhibit misalignment due to variations in installation, viewpoint, and focal length. We intentionally preserve this inherent misalignment. Furthermore, DVT-VOD1000 provides greater category diversity, which can enhance the model's generalization ability. As a drone-based platform, it also introduces significant challenges—such as fast motion, big scale variation, large rotation, and severe occlusion—that are less prevalent in fixed-viewpoint vehicle or surveillance systems, thereby demanding greater algorithmic robustness.

\begin{table*}[htbp]
\renewcommand{\arraystretch}{1.2}
  \centering
  \caption{We present a comprehensive comparison of our DVT-VOD1000 dataset with existing RGBT object detection datasets.}

    \begin{tabular}{lccccccc}
    \toprule    \textbf{Dataset}&\textbf{Type}&\textbf{Images}&\textbf{Resolution}&\textbf{Misaligned}&\textbf{Classes}&\textbf{Platform}&\textbf{Year}\\
    \midrule
    VEDAI \cite{razakarivony2016vehicle}&Image&12000&1024$\times$1024&\ding{53}&9&drone&2015\\
    KAIST \cite{choi2018kaist}&Image&95328&640$\times$480&\ding{53}&3&driving&2015\\
    CVC14 \cite{CVC14}&Image&17036&640$\times$512&\checkmark&1&driving&2016\\
    FLIR \cite{gonzalez2016pedestrian}&Image&24680&640$\times$512&\checkmark&4&driving&2019\\
    LLVIP \cite{jia2021llvip}&Image&30976&1080$\times$720&\ding{53}&1&surveillance&2021\\
    DroneVehicle \cite{sun2022drone}&Image&56878&640$\times$512&\ding{53}&5&drone&2022\\
    M$^3$FD \cite{liu2022target}&Image&8400&1024$\times$768&\ding{53}&6&various&2022\\
    DVTOD \cite{song2024misaligned}&Image&4358&1920$\times$1080&\checkmark&3&drone&2024\\
    VT-VOD50 \cite{wang2025erasure}&Video&18898&640$\times$368-1920$\times$1080&\ding{53}&7&traffic&2025\\
    DVT-VOD1000 (Ours)&Video&206928&960$\times$770&\Checkmark &15&drone&-\\
    \bottomrule
    \end{tabular}
  \label{dataset_compara}
\end{table*}%

\section{Experiments}
\begin{table*}[t]
	\centering
	\caption{
		We evaluate DCHNet alongside current mainstream detection models on DVT-VOD1000, emphasizing the best results in \textbf{bold}. The symbol ``-'' indicates that the measurement conditions are not met or that the result remains unobtainable. 
	}
    
		\begin{tabular}{lcc|cc|ccc}
			\toprule
			 Methods&  Backbone    &Type& AP50(\%) & AP(\%) &      FPS &   Params(M)    & FLOPs(G) \\
			\midrule
            Faster R-CNN \cite{ren2016faster}&ResNet-50&\multirow{13}{*}{Image}&22.8&8.5&26&41.4&134.4\\
            CFT \cite{qingyun2021cross} & CFB &    &11.9&3.98& \textbf{222.2} & 73.7 & - \\
			YOLOV10-M \cite{wang2024yolov10}&CSPNet&&28.3&8.91&210&16.5&64\\
            YOLOV13-S \cite{wang2024yolov10}&-&&28.4&9.26&476&9.0&\textbf{21}\\
            TOOD \cite{feng2021tood} &ResNet-50&&25&10.1&25.8&32&199\\
			Deformable DETR \cite{zhu2021deformable}&ResNet-50&&18.3&0.7&20.7&41.1&197\\
            RetinaNet \cite{lin2017focal} &ResNet-50&&22.6&9.2&38.4&36.62&168\\
            DINO \cite{zhang2022dino}&ResNet-50&&29.9&12&16.7&47.7&274\\
            DDQ DETR \cite{zhang2023dense} & ResNet-50 &       &29.4&12& 13 & 48.3& 275 \\
            $\mathrm{EI}^2\mathrm{Det}$ \cite{hu2025ei}& CSPDarknet53 &  & 30.2 & 12.4 & - & 128.66 &- \\
            RGFNet\cite{zhao2025reflectance}  & CSPDarknet &   &29.3&12.4 & 37.57 & 79.54 & 273 \\
            M-SpecGene \cite{zhou2025m} & ViT-B &   &29.5&11.8& 12.5 & 116& 538 \\
            Fusion-Mamba \cite{dong2025fusion} & YOLOv5 &   &29.4&11.4& 24.63 & 59.6& 112.9 \\
			\midrule
            DFF \cite{zhu2017deep}  & ResNet-50 & \multirow{6}{*}{Video}  &22.4&- & 40.4 & 62.1 & 24.9 \\
			
			FGFA \cite{zhu2017flow}  & ResNet-50 &    &23.7&-& 9 & 64.5& 41 \\
			
			RDN \cite{deng2019relation} & ResNet-50 &     &24.8&-& 11.3 &    -   & - \\

			MEGA \cite{chen2020memory}& ResNet-50 &    &25.0&-& 16.2 &   -    & - \\
			EINet \cite{wang2025erasure}  & Darknet53 &    &21.9&7.99& 204.2 & 11.6& 78.2 \\
			PTMNet \cite{wang2025high}  & CSPDarknet53 &   &23.0&8.82& 72.5 & 11.4& 77.8 \\
			\midrule
			DCHNet-S (Ours) & CSPDarknet53 & \multirow{2}{*}{Video}  & 28.5& 11.4 & 73 &\textbf{7.18}&27.2 \\
            DCHNet-L (Ours) & CSPDarknet53 &  & \textbf{31.7}& \textbf{13.7} & 23.3 & 82 &552.6 \\
			\bottomrule
	\end{tabular}
	\label{compara1}
\end{table*}
\subsection{Datasets}
We assess model performance on the existing VT-VOD50 dataset \cite{wang2025erasure} and our proposed DVT-VOD1000. The VT-VOD50 dataset, being the first dedicated to RGBT-VOD, features seven object categories across diverse real-world conditions like daytime and nighttime. This diversity allows it to effectively test algorithm robustness against significant illumination changes.
\subsection{Experimental Details}
We implement our experiments using the PyTorch framework with Python 3.8. The training runs on two NVIDIA GeForce RTX 3090 GPUs (each with 24GB memory), with DCHNet trained for 100 epochs at a batch size of 18 per GPU. The input resolution is set to 640$\times$640 pixels. To maintain spatial consistency across multimodal images, we disable potentially disruptive augmentations such as mosaic and mixup, retaining only color augmentation and horizontal flipping. All other training settings, including the learning rate optimizer, remain consistent with the baseline. For image-based and unimodal detectors, pixel-level fusion of RGB and thermal images is performed to accommodate multimodal inputs.

We primarily evaluate and compare the model's detection accuracy and inference speed. For accuracy assessment, we employ the mAP and AP50 metrics, which are standard in object detection tasks. The Average Precision (AP) measures the detection performance for a single object class; it is characterized by its ability to balance precision and recall across various confidence thresholds. AP50 specifically denotes the AP calculated at an IoU threshold of 0.5. In multi-class detection scenarios, the mean Average Precision (mAP) represents the mean of AP values across all classes, thereby indicating the model's overall detection capability on the dataset. For speed evaluation, we use FPS, a common metric in video analysis that quantifies the number of frames processed per second and indicates whether the model meets real-time processing requirements.

\subsection{Comparative Experiments}
To comprehensively compare the performance of our DCHNet with existing mainstream detectors, a thorough evaluation is conducted on two RGBT VOD datasets: VT-VOD50 and the proposed DVT-VOD1000. Our DCHNet is available in two versions: DCHNet-S, which is designed to balance accuracy and efficiency, and DCHNet-L, which is optimized for maximum performance. DCHNet-L is a scaled‑up version of DCHNet-S with increased depth and width.

\subsubsection{Results in DVT-VOD1000}
Table \ref{compara1} presents the results of training and evaluating various detectors on the proposed DVT-VOD1000 dataset. Our DCHNet-L achieves an AP50 score of 31.7\%, ranking first among all compared methods and outperforming the second-best method, $\mathrm{EI}^2\mathrm{Det}$ (30.2\%), by 1.5 \%. Our DCHNet-S achieves an AP50 score of 28.5\%, surpassing all other video detectors and exceeding the second-best video-based method, MEGA (25.0\% AP50), by 3.5\%. DCHNet-S also demonstrates superior performance compared to most image-based detectors. Moreover, DCHNet-S strikes a favorable balance between model efficiency and accuracy, delivering 28.5\% AP50 at 73 FPS with only 7.18M parameters and 27.2G FLOPs. In contrast, $\mathrm{EI}^2\mathrm{Det}$ has a substantially larger parameter count of 128.66M. Although EINet and PTMNet also maintain relatively small parameter sizes, their AP50 scores are only 21.9\% and 23.0\%, respectively—considerably lower than that of DCHNet-S.

\begin{table}[t]
	\centering
	\caption{
		We evaluate DCHNet alongside current mainstream detection models on VT-VOD50, emphasizing the best results in \textbf{bold}. The symbol ``-'' indicates that the measurement conditions are not met or that the result remains unobtainable. 
	}\resizebox{\linewidth}{!}{
		\begin{tabular}{lcc|cc}
			\toprule
			Methods&  Backbone    &Type  & AP50(\%) & AP(\%)  \\
			\midrule
			YOLOV3 \cite{redmon2018yolov3}& Darknet53 & Image     &33.9& 17.4\\
			
			CFT \cite{qingyun2021cross} & CFB & Image   &42.5&18.9\\
	
			YOLOV9-C \cite{wang2024yolov9}&GELAN&Imgae&49.1&26.9\\
			
			YOLOV10-M \cite{wang2024yolov10}&CSPNet&Image&46.2&25.2\\
			
			TOOD \cite{feng2021tood} &ResNet-50&Image&36.3&19\\
			
			Deformable DETR \cite{zhu2021deformable}&ResNet-50&Imgae&42.5&23.3\\
			
			RT-DETR \cite{zhao2024detrs}&ResNet-50&Imgae&40.2&21.6\\
			
			DINO \cite{zhang2022dino}&ResNet-50&Image&47.4&25.9\\

			DDQ DETR \cite{zhang2023dense} & ResNet-50 & Image &48.3&26.5\\
			
			DiffusionDet \cite{chen2023diffusiondet}&ResNet50&Image&46.9&25.1\\
            
            $\mathrm{EI}^2\mathrm{Det}$ \cite{hu2025ei}& CSPDarknet53 &Image  & 53.6 & 26.3\\
            RGFNet\cite{zhao2025reflectance}&CSPDarknet&Image&52.9&29.3\\
            
            M-SpecGene \cite{zhou2025m} & ViT-B & Image  &49.6&29.2\\
            Fusion-Mamba \cite{dong2025fusion} & YOLOv5 &Image & 56&30\\
            
			\midrule
			DFF \cite{zhu2017deep}  & ResNet-50 & Video  &33.5&14.1 \\
			
			FGFA \cite{zhu2017flow}  & ResNet-50 & Video   &35.1&15.8\\
			
			RDN \cite{deng2019relation} & ResNet-50 & Video    &40&- \\
			
			SELSA \cite{wu2019sequence}  & ResNet-50 & Video  &39.4&17.4\\
			
			MEGA \cite{chen2020memory}& ResNet-50 & Video   &27.8&-\\
			
			Temporal ROI Align \cite{gong2021temporal} & ResNet-50 & Video   &38&17\\
			
			CVA-Net \cite{10.1007/978-3-031-16437-8_59}  & ResNet-50 & Video &39.7&19.7 \\
			
			STNet \cite{qin2023spatial}  & ResNet-50 & Video  &38.4&18.4\\
			
			EINet \cite{wang2025erasure}  & Darknet53 & Video   &46.3&24\\
			PTMNet \cite{wang2025high}  & CSPDarknet53 & Video  &51.4&27.1\\
			MSENet \cite{wang2025mixturescaleexpertsalignmentfree}  & CSPDarknet53 & Video  &50.3&27.6\\
			\midrule
			DCHNet-S (Ours) & CSPDarknet53 &Video  & 55.5& 30\\
            DCHNet-L (Ours) & CSPDarknet53 &Video  & \textbf{57.5}& \textbf{33.7}\\
			\bottomrule
	\end{tabular}}
	\label{compara2}
\end{table}

Table \ref{compara1} further reveals that the AP50 scores of most video object detectors remain below 25\%, significantly lagging behind high-performing multimodal image detectors such as $\mathrm{EI}^2\mathrm{Det}$ and M-SpecGene. This suggests that temporal information aggregation alone offers limited benefits, whereas a well-designed cross-modal fusion mechanism plays a more critical role in enhancing detection performance in RGBT videos.

\subsubsection{Results in VT-VOD50}
Table \ref{compara2} presents the results of training and evaluating various detectors on the VT-VOD50 dataset. Even powerful image-based detectors, such as DDQ DETR (48.3\% on AP50) and YOLOV9-C (49.1\% on AP50), still underperform dedicated video-based detectors like MSENet and PTMNet when applied to video streams, which often contain motion blur and occlusion. In contrast, our proposed DCHNet-L achieves 57.5\% on AP50, surpassing the YOLOV9-C by 8.4\%. RGFNet \cite{zhao2025reflectance} is specifically designed to address spatial misalignment in RGB-thermal image pairs. Despite not leveraging temporal information, it achieves a strong performance of 52.9\% AP50. $\mathrm{EI}^2\mathrm{Det}$ also achieves excellent detection performance, benefiting from the guidance of illumination and edge information. Fusion-Mamba achieves the best performance among the compared image detectors, reaching 56\% AP50 through cross-modal interactions in the hidden state space. Among the video detectors evaluated, MSENet \cite{wang2025mixturescaleexpertsalignmentfree} and PTMNet \cite{wang2025high} achieve the great performance, both featuring dedicated multimodal fusion designs. This underscores that video detectors designed for single-modality input (such as CVA-Net and STNet) are constrained by the limitations of a single imaging modality and thus fail to fully exploit the complementary information from RGB and thermal images. Our DCHNet-S and DCHNet-L outperform PTMNet by 4.1\% and 6.1\% on AP50, respectively. This significant improvement demonstrates the effectiveness of our fusion strategy in leveraging multimodal temporal information.

\subsection{Ablation Studies}
We conduct detailed ablation studies on each component of DCHNet to validate the effectiveness of the proposed method.
\subsubsection{Study on Each Component}
To validate the effectiveness of our proposed modules, we incrementally construct DCHNet from the baseline method, as illustrated in Table \ref{abla}. Groups (a) and (b) present the training and testing results on the baseline using RGB and thermal images, respectively. It is observed that using only thermal images as input yields significantly inferior performance compared to using RGB ones, as thermal images not only lack crucial texture and detail information but also exhibit spatial misalignment with the RGB-based ground-truth annotations. This further underscores the necessity of a dedicated RGBT fusion strategy. The group (c) reports the results of applying PSAM for alignment and information aggregation using two temporal RGB frames, achieving a 1.6\% improvement on AP50 over (a). The group (d) shows the results of employing PSAM for spatial alignment and feature fusion with one RGB and one thermal image as input, leading to gains of 1.7\% and 16.9\% over (a) and (b), respectively, demonstrating the effectiveness of PSAM in fusing spatially unaligned multimodal inputs. The group (e) presents the results obtained by incorporating LBP to inject RGB edge information into the thermal modality, building upon the configuration in (d). Finally, by integrating PSAM, LBP, and DHFM, we construct DCHNet, which achieves an overall AP50 improvement of 2.9\% over the group (a).

\begin{table}[t]
	\centering
	\caption{Ablation study on each component of DCHNet. The study is based on the DVT-VOD1000 dataset.
	}
    \resizebox{\linewidth}{!}{
		\begin{tabular}{lccccccccc}
			\toprule
			Groups&Data&PSAM&LBP&DHFM& AP50(\%) & AP(\%) &Params(M)& FLOPs(G) \\
			\midrule
			(a)&RGB&&&&25.6&9.9&3.01&8.21\\
            (b)&T&&&&10.4&4.1&3.01&8.21\\
            (c)&RGB&\checkmark&&&27.2&11&4.43&15.9\\
            (d)&RGBT&\checkmark&&&27.3&11.3&4.51&11.8\\
            (e)&RGBT&\checkmark&\checkmark&&28&11.5&5.8&21.8\\
            DCHNet&RGBT&\checkmark&\checkmark&\checkmark&28.5&11.4&7.18&27.2\\
			\bottomrule
	\end{tabular}}
	\label{abla}
\end{table}

\subsubsection{Study on Patch Size in PSAM}
To further investigate the alignment mechanism of PSAM, we conduct a series of ablation experiments focusing on patch size. The results in Table \ref{patchsize} show that a $5\times5$ patch size achieves the highest accuracy, improving AP50 by 1.0\% and 1.3\% compared to the $8\times8$ and $3\times3$ patch sizes, respectively. Our analysis indicates that a $3\times3$ patch size is too small to capture sufficiently discriminative spatial context features. In contrast, an $8\times8$ patch size introduces excessive background and redundant information, which may complicate the alignment process and ultimately degrade accuracy.
\begin{table}[t]
	\centering
	\caption{ Ablation study on patch size in PSAM. The study is based on the DVT-VOD1000 dataset.
	}
    \resizebox{\linewidth}{!}{
		\begin{tabular}{lccccc}
			\toprule
			Groups& patch size&  AP50(\%) & AP(\%) &Params(M)   & FLOPs(G) \\
			\midrule
			(a) (Ours) &5$\times$5&27.2&11&7.18&27.2\\
            (b)&8$\times$8&26.2&10.6&7.31&27.4\\
            (c)&3$\times$3&25.9&10.1&7.02&27.1\\
			\bottomrule
	\end{tabular}}
	\label{patchsize}
\end{table}
\begin{table}[t]
	\centering
	\caption{ Ablation Study on RGBT fusion methods. The study is based on the VT-VOD50 dataset.
	}
		\begin{tabular}{lcccc}
			\toprule
			Methods& AP50(\%) & AP(\%) &Params(M)   & FLOPs(G) \\
			\midrule
			Cross Attention\cite{lin2022cat}&53.4&28.8&7.37 &27.4\\
            DCEvo\cite{liu2025dcevo} &54.5 & 29.2&8.8&32.6\\
            TDF\cite{wang2025high}&49.6&26.9&20.9&32.4\\
            TDFusion\cite{bai2025task}&37.7&20.4&8.38&31.5\\
            DHFM (Ours) &55.5&30&7.18&27.2\\
			\bottomrule
	\end{tabular}
	\label{comrgbt}
\end{table}
\subsubsection{Comparison of DHFM with Other Fusion Methods}
We conduct a series of ablation experiments on the RGBT fusion method to demonstrate the effectiveness of our DHFM, with the results presented in Table \ref{comrgbt}. The results demonstrate that our DHFM achieves an AP50 of 55.5\%, which is 1.0\% higher than the second-best method, DCEvo (54.5\%), and an AP of 30.0\%, also surpassing DCEvo (29.2\%). Compared to Cross Attention, DHFM outperforms it by 2.1\% in AP50, highlighting its effectiveness in cross-modal fusion. Moreover, DHFM has only 7.18 million parameters—the smallest among all compared methods—along with the lowest computational cost, indicating that DHFM strikes a favorable balance between efficiency and accuracy. In contrast, TDF and TDFusion yield lower accuracy, with the latter not undergoing task-specific fine-tuning, which may account for its suboptimal performance.

\section{Conclusion}
This work proposes a novel Dual-Correlation Hypergraph Network (DCHNet) for RGBT VOD, which alleviates the challenge of cross-modal spatial misalignment. We first propose PSAM to perform spatial transformation and alignment of different local regions in RGBT image pairs, as the misalignment varies across different locations in the image. We then propose DHFM, which leverages a hypergraph to model high-dimensional feature relationships between RGB and T, thereby aggregating complementary features across modalities more effectively. Additionally, we construct a large-scale dataset, named DVT-VOD1000, to evaluate RGBT VOD methods. This dataset is collected from a drone platform and consists of 1000 videos spanning 15 common categories. DVT-VOD1000 covers a wide range of real-world scenes, providing comprehensive evaluation capabilities and offering significant potential for RGBT VOD research.


\bibliographystyle{IEEEtran}
\bibliography{IEEEabrv,mylib}

\end{document}